\documentclass[letterpaper, 10 pt, conference]{ieeeconf}  

\IEEEoverridecommandlockouts                              
\usepackage{amsmath} 
\usepackage{amssymb}  
\usepackage{cite}
\usepackage{threeparttable}
\usepackage{stfloats}
\usepackage{graphicx}\usepackage{multirow}
\usepackage{url}
\usepackage{algorithm}
\usepackage{algorithmic}
\usepackage{subcaption} 
\usepackage{comment}
\allowdisplaybreaks

\title{\LARGE \bf Adaptive Probabilistic Vehicle Trajectory Prediction Through Physically Feasible Bayesian Recurrent Neural Network}
\author{Chen Tang, Jianyu Chen and Masayoshi Tomizuka
\thanks{{$^\dagger$}This work was supported by DENSO International America, Inc}
\thanks{{$^\dagger$}C. Tang, J. Chen and M. Tomizuka are with the Department of Mechanical
Engineering, University of California, Berkeley, CA 94720 USA (e-mail:
chen\_tang@berkeley.edu, jianyuchen@berkeley.edu, tomizuka@berkeley.edu).}
}

\begin{document}

\maketitle
\thispagestyle{empty}
\pagestyle{empty}

\begin{abstract}
Probabilistic vehicle trajectory prediction is essential for robust safety of autonomous driving. Current methods for long-term trajectory prediction cannot guarantee the physical feasibility of predicted distribution. Moreover, their models cannot adapt to the driving policy of the predicted target human driver. In this work, we propose to overcome these two shortcomings by a Bayesian recurrent neural network model consisting of Bayesian-neural-network-based policy model and known physical model of the scenario. Bayesian neural network can ensemble complicated output distribution, enabling rich family of trajectory distribution. The embedded physical model ensures feasibility of the distribution. Moreover, the adopted gradient-based training method allows direct optimization for better performance in long prediction horizon. Furthermore, a particle-filter-based parameter adaptation algorithm is designed to adapt the policy Bayesian neural network to the predicted target online. Effectiveness of the proposed methods is verified with a toy example with multi-modal stochastic feedback gain and naturalistic car following data. 
\end{abstract}

\section{Introduction}
Safety is critical for autonomous vehicle (AV), because it needs to interact with the complicated driving environment filled with human-driven vehicles and pedestrians. It is then necessary for AV to predict other vehicles' future trajectories accurately. Many efforts have been made to enhance prediction accuracy \cite{lefevre2014survey, houenou2013vehicle, liu2016enabling, lin2018deep, jeong2017long}. However, deterministic prediction is not sufficient to enable robust safety because human driver's decisions are uncertain even under the same circumstances. Instead, the probabilistic distribution of trajectories should be estimated. 

Various methods exist in literature for probabilistic trajectory prediction. Physics-based motion models are developed based on vehicle dynamics or kinematic model. Certain assumption on control inputs is made to obtain a self-evolving model, for instance, constant turn rate and velocity (CTRV) model \cite{polychronopoulos2007sensor}. Stochastic trajectory prediction can be achieved through Bayesian filtering \cite{veeraraghavan2006deterministic} or Monte Carlo simulation \cite{eidehall2008statistical,althoff2011comparison}. The predicted distribution satisfies certain physical constraints imposed by the vehicle model. However, they are precise only for short time horizon \cite{ortiz2011behavior}. Another stream of works skips the physics and directly constructs a probabilistic model of trajectory. Commonly used models include Gaussian mixture model (GMM) \cite{wiest2012probabilistic} and Gaussian Process (GP) \cite{joseph2011bayesian}. Moreover, the prevalence of deep learning facilitated the usage of deep recurrent neural networks (RNN) in probabilistic trajectory prediction \cite{kim2017probabilistic, park2018sequence}. This category of methods have better performance for longer prediction horizon but feasibility is not guaranteed. In practice, they are usually combined with maneuver recognition modeule, resulting in a hierarchical structure \cite{xie2018vehicle}. However, human driver's intention in complicated scenarios can hardly be labeled, preventing accurate training of the maneuver recognition module. 

To enhance robustness, the models need to be trained with trajectories executed by different human drivers to avoid overfitting with specific driving strategies. On the other hand, when deploying the trained models, the specific target driver should have relatively consistent characteristics over time. Therefore, its policy should have smaller variance compared to the distribution in the training dataset. Consequently, accuracy could be increased if we are able to adapt the model to the specific target of interest online. This issue is considered in \cite{woo2018trajectory}. Individual driving characteristics are encoded as parameters in potential field functions, which are identified online. However, its performance is restricted by its relatively simple and inflexible model structure. Moreover, it is designed for a particular scenario without the capability for generalization. 

In this work, we aim to combine the advantages of both categories of methods. Concretely, our objective is to design a probabilistic model predicting dynamically feasible trajectory distribution for long horizon. More importantly, we want to make it adaptable to the specific target vehicle online. We divide the entire model into two components, policy and dynamics. The policy describes how the agent reacts to his observations. The dynamics describes how the scenario propagates given his actions. Dynamics model at certain level of complexity can normally be assumed known for vehicles \cite{schubert2008comparison}. It can then be combined with an expressive data-driven model representing the complicated intrinsic policy, forming an unit model for a single time step. To infer trajectory distribution, units for the entire prediction horizon are then connected into a recurrent structure. In this case, the problem is transformed into a probabilistic policy learning problem, which is similar to the idea of probabilistic model-based imitation learning \cite{englert2013probabilistic}. 

We propose to model the policy with Bayesian neural network (BNN) which enables complicated output distribution, e.g. multi-modal distribution. It can therefore unify the commonly used heuristic-based hierarchical model structure, so that the policy model is optimized as a whole. Potentially, it can be connected with deep-learning-based perception and decision making modules. Bayesian deep learning can be applied in an end-to-end fashion to optimize the entire architecture \cite{mcallister2017concrete}. Its flexibility in distribution representation also enables a policy model for general driving scenarios, where a single parametric family of distribution is not sufficient. We adopt a gradient-based method black-box $\alpha$-divergence minimization (BB-$\alpha$) \cite{hernandez2016black} to train the entire recurrent model for better long-term prediction performance. Eventually, we obtain a Bayesian recurrent neural network (BRNN). To achieve adaptation, we propose a particle-filter-based online adaptation method to incorporate new samples quickly online. The model after training is treated as a prior to obtain the posterior policy BNN adapted to the target human driver. 

Contribution of this work is threefold: 1) We propose to convert the trajectory prediction problem into a policy learning problem and develop a BRNN model for long-term trajectory prediction. (2) Particle-filter-based online adaptation method is proposed to adapt the policy BNN model online to the specific predicted target. (3) The proposed methods are applied to solve a practical vehicle trajectory prediction problem in car following scenario. 

\section{Background} \label{background}
\subsection{Bayesian Neural Networks with Stochastic Inputs}
A BNN with random input noise $z$ is developed in \cite{depeweg2016learning} for stochastic dynamical system modeling. Given data $\mathcal{D}=\{\boldsymbol{x}_n, \boldsymbol{y}_n\}_{n=1}^{N}$, where $\boldsymbol{x}_n$ is feature vector and $\boldsymbol{y}_n$ is labeled output, it is assumed that $\boldsymbol{y}_{n}=g\left(\boldsymbol{x}_{n}, z_n;\mathcal{W}\right)+{\epsilon}_n$, where $g\left(\cdot,\cdot;\mathcal{W}\right)$ represents a neural network with $L$ layers and $L_l$ hidden units in layer $l$, parameterized by $\mathcal{W}=\{\boldsymbol{W}_l\}_{l=1}^L$, a collection of $V_l\times\left(V_{l-1}-1\right)$ weight matrices. The elements of $\boldsymbol{W}_l$ are mutually independent and an identical Gaussian prior is assigned to each of them, i.e. $w_{ij,l}\sim\mathcal{N}(0,\lambda)$. $z_n$ is a synthetic stochastic disturbance to enable richer probabilistic distribution structure. Each pair of sample $(\boldsymbol{x}_n, \boldsymbol{y}_n)$ has a corresponding independent $z_n$. $\boldsymbol{\epsilon}_i$ is random noise to enable continuous distribution of $\boldsymbol{y}_{n}$. Both $z_n$ and ${\epsilon}_i$ have Gaussian priors, i.e. $z_n\sim\mathcal{N}(0,\gamma)$ and ${\epsilon}_n\sim\mathcal{N}(0,\Sigma_{{\epsilon}})$. The model is trained by estimating the posterior distribution of parameters $p\left(\mathcal{W},\boldsymbol{z}|\mathcal{D}\right)$, where $\boldsymbol{z}$ is the vector collecting all the random inputs. The distribution $p\left(\mathcal{W},\boldsymbol{z}|\mathcal{D}\right)$ is approximated by $q\left(\mathcal{W},\boldsymbol{z}\right)$ as in (\ref{eqn:q}). The conditional output distribution $p(\boldsymbol{y}|\boldsymbol{x},\mathcal{D})$ can then be approximated by Monte Carlo sampling, i.e. computing samples of $\boldsymbol{y}$ with samples of $\mathcal{W}, \boldsymbol{z}$ drawn from $q(\mathcal{W},\boldsymbol{z})$. The structural complexity and non-linearity embedded in the neural network enable abundant probabilistic structure, strengthening the expressiveness of the model.
\begin{equation} \label{eqn:q}
\begin{aligned}
&q\left(\mathcal{W},\boldsymbol{z}\right)\\
=&\prod_{l=1}^{L}\prod_{i=1}^{V_l}\prod_{j=1}^{V_{l-1}+1}\mathcal{N}\left(w_{ij,l}|m_{ij,l}^{w},v_{ij,l}^{w}\right)\prod_{n=1}^N\mathcal{N}(z_n|m_n^z,v_n^z)  \\
\propto&\left[\prod_{n=1}^{N}f(\mathcal{W})f_n(z_n)\right] p(\mathcal{W})p(\boldsymbol{z}),
\end{aligned}
\end{equation}
where
\begin{equation*}
\begin{aligned}
p(\mathcal{W})&=\prod_{l=1}^{L}\prod_{i=1}^{V_l}\prod_{j=1}^{V_{l-1}+1}\mathcal{N}(w_{ij,l}|0,\lambda)\\ p(\boldsymbol{z})&=\prod_{n=1}^N\mathcal{N}(z_n|0,\gamma)\\
f_n(z_n)&=\exp \left\{\frac{v_n^z-\gamma}{2\gamma v_n^z}z_n^2+\frac{m_n^z}{v_n^z}z_n\right\}\\
f(\mathcal{W})&=\\
\exp&\left\{\sum_{l=1}^{L}\sum_{i=1}^{V_l}\sum_{j=1}^{V_{l-1}+1}\frac{1}{N}\left(\frac{v_{ij,l}^w-\lambda}{2\lambda v_{ij,l}^w}w^2_{ij,l}+\frac{m_{ij,l}^w}{v_{ij,l}^w}w_{ij,l}\right)\right\}\\
\end{aligned}
\end{equation*}
\subsection{Black-Box $\alpha$-Divergence Minimization}
BB-$\alpha$ \cite{hernandez2016black} can be used to approximate the distribution $q(\mathcal{W},\boldsymbol{z})$, which is a gradient-based and sampling-based approximate inference method. At each iteration, M samples of $\mathcal{W}$ and $\boldsymbol{z}$ are drawn from current estimated posterior distribution. A mini-batch of data $\mathcal{S}\subset \{1, 2, 3,..., N\}$ is sampled uniformly. The noisy estimate of energy function as shown in (\ref{eqn:energy}) is minimized by tuning $q(\mathcal{W},\boldsymbol{z})$ and $\Sigma$ through gradient descent. The only term related to the data is the likelihood function $p\left(\boldsymbol{y}_n|\boldsymbol{x}_{n}, \mathcal{W}_{s}, z_{n,s}, \Sigma \right)$. One advantage of BB-$\alpha$ is that the divergence function used in the energy function can be tuned by adjusting $\alpha$, providing flexibility to improve performance \cite{hernandez2016black}. A general guidance for $\alpha$ value selection is provided in \cite{depeweg2016learning}. In practice, the best $\alpha$ value can be directly found by validation. 

\begin{equation} \label{eqn:energy}
\hat{E}_\alpha(q)=-\log{Z_q} -\frac{N}{\alpha|\mathcal{S}|}\sum_{n\in{\mathcal{S}}}\hat{L}^n_\alpha(q),\\
\end{equation}
where
\begin{equation*}
\begin{aligned}
\log Z_q &= \sum_{l=1}^{L}\sum_{i=1}^{V_l}\sum_{j=1}^{V_{l-1}+1}\left(\frac{1}{2}\log(2\pi v^w_{ij,l})+\frac{(m_{ij,l}^w)^2}{2v_{ij,l}^w}\right)\\
+ \sum_{n=1}^N & \left[\frac{1}{2}\log(2\pi v_n^z)+\frac{(m_n^z)^2}{2v_n^z}\right]\\
\hat{L}_{\alpha}^{n}(q)&=
\log\frac{1}{M}\sum_{s=1}^{M}\left(\frac{p\left(\boldsymbol{y}_n|\boldsymbol{x}_{n}, \mathcal{W}_{s}, z_{n,s}, \Sigma \right)}{f(\mathcal{W}_s)f(z_{n,s})}\right)^\alpha.
\end{aligned}
\end{equation*}

\section{Problem Formulation} \label{problem}
\begin{figure}[t]
\centering
\includegraphics[scale=0.24]{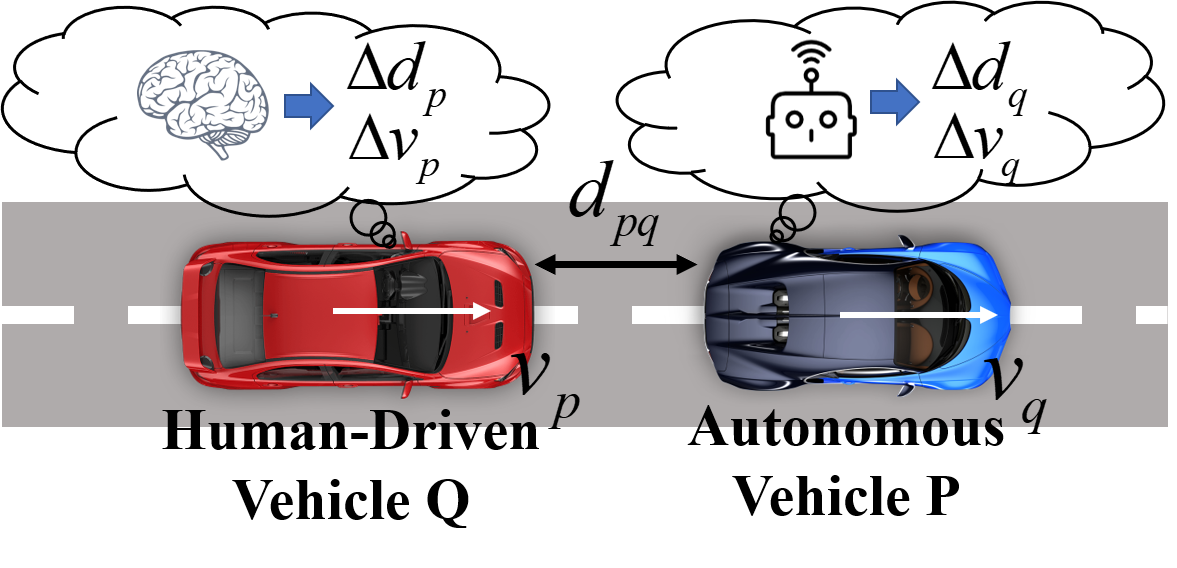}
\caption{Interactive Car Following Scenario with human-driven vehicle $Q$ and autonomous vehicle $P$. } 
\label{fig:scenario}
\end{figure}

In this section, we formulate the trajectory prediction problem and introduce the notations we use in latter sections. The problem is formulated in the context of the interactive car following scenario that our proposed methods are verified with. It can be generalized to other scenarios and other trajectory prediction problems easily. The following vehicle, denoted by $P$, is driven by human driver and the preceding vehicle, denoted by $Q$, is an AV. We define a state variable $\boldsymbol{x}=[d_{pq}\ v_p\ v_q]^{\rm T}$. $d_{pq}$ is the longitudinal distance between the two vehicles. $v_p$ and $v_q$ are the longitudinal speed of $P$ and $Q$. The actions executed by the two vehicles are denoted as $\boldsymbol{a}_p=[\Delta{d}_p\ \Delta{v}_p]^{\rm T}$ and $\boldsymbol{a}_q=[\Delta{d}_q\ \Delta{v}_q]^{\rm T}$. $\Delta{d}_p$ and $\Delta{d}_q$ are the longitudinal displacement of $P$ and $Q$ over the sampling time $\Delta t$. $\Delta{v}_p$ and $\Delta{v}_q$ are the change of longitudinal speed of $P$ and $Q$ over $\Delta t$. The scenario is modeled as a discrete-time system whose dynamics is described below and illustrated in Figure $\ref{fig:scenario}$. 
\begin{align} 
d_{pq,i+1}&=d_{pq,i}-\Delta{d}_{p,i}+\Delta{d}_{q,i} \label{eqn:car_model1}+\omega_{1,i}\\
v_{p,i+1}&=v_{p,i}+\Delta{v}_{p,i}+\omega_{2,i} \label{eqn:car_model2}\\
v_{q,i+1}&=v_{q,i}+\Delta{v}_{q,i}+\omega_{3,i} \label{eqn:car_model3}.
\end{align}
In the equations, $\omega_{1,i}$, $\omega_{2,i}$ and $\omega_{3,i}$ are additive Gaussian noise accounting for the uncertainty in dynamics. We denote ${\omega}_i \sim \mathcal{N}(0,\Sigma_{{\omega_i}})$ as the random vector collecting these three noise variables. For convenience, the dynamics model is denoted as the following general form in the remainder:
\begin{equation} \label{eqn:model}
\boldsymbol{x}_{i+1}=h\left(\boldsymbol{x}_{i}, \boldsymbol{a}_{p,i}, \boldsymbol{a}_{q,i}\right)+{\omega_i}
\end{equation}

we assume we know $\boldsymbol{x}$ for the current and past time steps and the planned future actions of the AV, $\boldsymbol{a}_q$, for next $h$ time steps. At a time step $k$, given a trajectory of $\boldsymbol{x}$ from time step $k-m$ to $k$, $\boldsymbol{x}_{k-m:k}$, and a series of $\boldsymbol{a}_q$ from time step $k$ to $k+h-1$, $\boldsymbol{a}_{k:k+h-1}$,the conditional distribution of future state trajectory $p\left(\boldsymbol{x}_{k+1:k+h}|\boldsymbol{x}_{k-m:k}, \boldsymbol{a}_{q, k:k+h-1}\right)$ is predicted. The predicted distribution is utilized by sampling-based motion planning algorithms which consider interaction and uncertainty \cite{jianyu2018continuous}.

To fulfill the goal, we need to construct a probabilistic model of trajectory. In this work, we break down the probabilistic model into units connected recurrently. Each unit corresponds to a single time step, describing the transition of state at this time step. At a time step $i$, we assume that the state at next time step $\boldsymbol{x}_{i+1}$ depends on the states of previous $m+1$ time steps, i.e. $\boldsymbol{x}_{i-m:i}$. Consequently, each unit describes the probabilistic transition function $p_{\boldsymbol{\theta_i}}\left(\boldsymbol{x}_{i+1}|\boldsymbol{x}_{i-m:i},\boldsymbol{a}_{q,i}\right)$, parameterized by $\boldsymbol{\theta_i}$ for $i=k,...,k+h-1$. By assuming the unit is time-invariant, the transition functions at all time steps are parameterized by an identical parameter vector $\theta$. The probabilistic transition function can be further split up as follows:
\begin{equation} \label{eqn:transition}
\begin{aligned}
&p_{\boldsymbol{\theta}}\left(\boldsymbol{x}_{i+1}|\boldsymbol{x}_{i-m:i},\boldsymbol{a}_{q,i}\right) \\ =&\int{p\left(\boldsymbol{x}_{i+1}|\boldsymbol{x}_i,\boldsymbol{a}_{q,i},\boldsymbol{a}_{p,i}\right)p_{\boldsymbol{\theta}}\left(\boldsymbol{a}_{p,i}|\boldsymbol{x}_{i-m:i}\right)}d\boldsymbol{a}_{p,i}.
\end{aligned}
\end{equation}
The function $p_{\boldsymbol{\theta}}\left(\boldsymbol{a}_i|\boldsymbol{x}_{i-m:i}\right)$ represents $P$'s inherent policy. We assume that it depends only on $\boldsymbol{x}_{i-m:i}$, but not on $\boldsymbol{a}_{q,i}$. It is reasonable as human drivers make decisions according to their current and historical observations. Other drivers' decisions cannot be directly observed. $p\left(\boldsymbol{x}_{i+1}|\boldsymbol{x}_{i},\boldsymbol{a}_{p,i},\boldsymbol{a}_{q,i}\right)$ represents the dynamics. Since the dynamics model is known as (\ref{eqn:model}), we can obtain 
$p\left(\boldsymbol{x}_{i+1}|\boldsymbol{x}_{i},\boldsymbol{a}_{p,i},\boldsymbol{a}_{q,i}\right)
=\mathcal{N}\left(\boldsymbol{x}_{i+1}; h\left(\boldsymbol{x}_{i}, \boldsymbol{a}_{p,i}, \boldsymbol{a}_{q,i}\right), \Sigma_{\boldsymbol{\omega_i}}\right)$.
By contrast, probabilistic model of the policy is unknown and need to be identified, which is the key element in our proposed model.
\section{Methodology} \label{method}
In this work, we modify the BNN with random input noise introduced in previous section to model the stochastic control policy. To enhance prediction performance for long time horizon, the entire recurrent model consisting of policy BNN and physical model is trained via BPTT. Accordingly, the likelihood function in (\ref{eqn:energy}) is replaced by the likelihood of labeled future trajectory, estimated with BNN parameter sampling and optimal filtering. The parameters of the policy BNN are adapted online through particle filtering. Intuitively, weight samples generating trajectory similar to the actual trajectory are assigned higher weights. Consequently, the mass of weight distribution is shifted toward those weights. 
\subsection{Physically Feasible Bayesian Recurrent Neural Network}
We use the BNN with random input noise to represent the policy. Different from \cite{depeweg2016learning}, we assume that the random inputs for all the samples are identical, i.e. $z_i=z_2=\dots=z_N = z$. Otherwise, it is ambiguous which $z_n$ should be used for a test sample. Concretely, we assume that $\boldsymbol{a}_{p,i}=g\left(\boldsymbol{x}_{i-m:i}, z;\mathcal{W}\right)+{\epsilon}_i$ at each time step $i$. The posterior distribution is then approximated with (\ref{eqn:tilde_q}) instead.

\begin{figure}[t]
\centering
\includegraphics[scale=0.31]{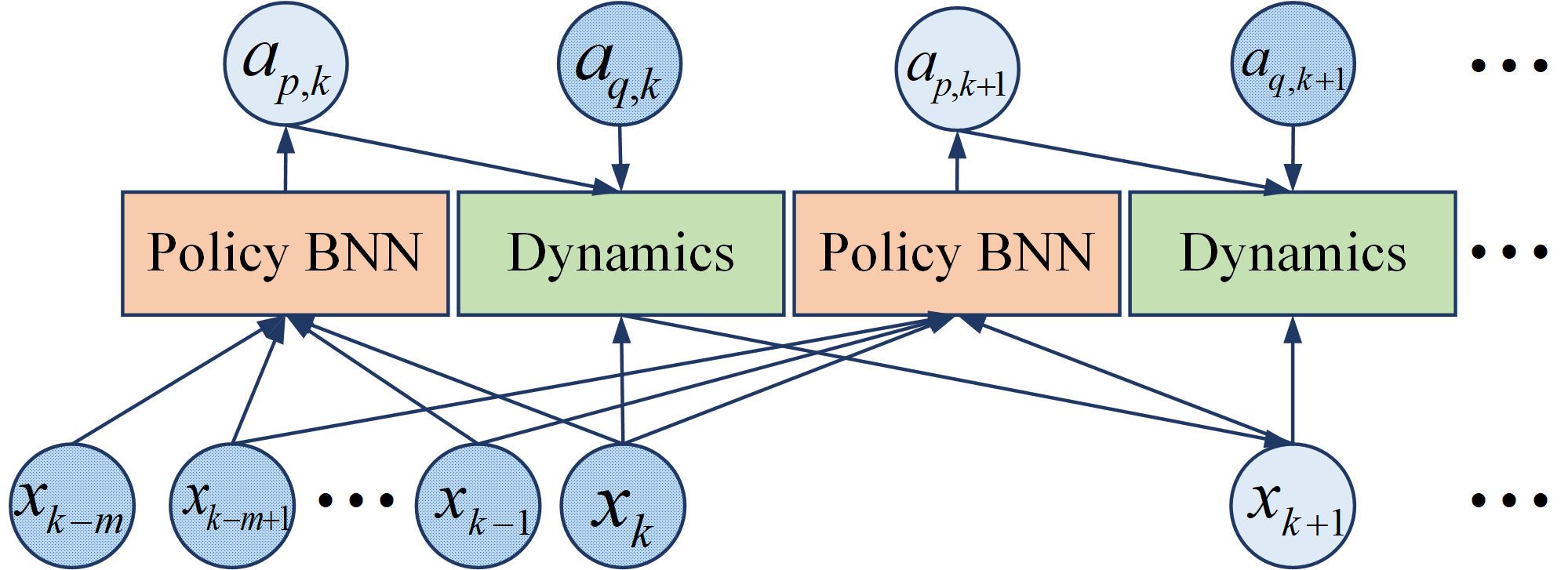}
\caption{Structure of the physically feasible BRNN. The shadowed nodes $\boldsymbol{x}_{k-m:k}$ and $\boldsymbol{a}_{q,k:k+h-1}$ are given.} 
\label{fig:structure}
\end{figure}

With the policy BNN, the recurrent model developed in last section becomes a BRNN with dynamics model embedded (Figure \ref{fig:structure}). Instead of training the policy BNN alone, we propose to directly train the entire recurrent model. Since BB-$\alpha$ is a gradient-based method, it is straightforward to extend the scheme to train the BRNN via BPTT \cite{werbos1990backpropagation,fortunato2017bayesian}. The training dataset consists of state trajectories, planned future input series and actual future trajectories, i.e. $\mathcal{D}=\{\boldsymbol{x}_{-m:0}^n, \boldsymbol{a}^n_{q,0:h-1},\boldsymbol{x}^n_{1:h}\}_{n=1}^{N}$. In addition to $\mathcal{W}$ and $\boldsymbol{z}$, samples of addictive noise series $\{{\omega}_{0:h-1}^{n}\}_{n=1}^{N}=\{\boldsymbol{\omega}^n\}_{n=1}^{N}$ and $\{{\epsilon}_{0:h-1}^{n}\}_{n=1}^{N}=\{\boldsymbol{\epsilon}^n\}_{n=1}^{N}$ are drawn at each iteration. The energy function is essentially the same. The major difference is that the likelihood function becomes $p\left(\boldsymbol{x}_{1:h}^{n}|\boldsymbol{x}_{-m:0}^n, \boldsymbol{a}_{q,0:h-1}^n, \mathcal{W}_{s}, z_s, \boldsymbol{\epsilon}_s^n, \boldsymbol{\omega}_s^n \right)$, namely the likelihood of labeled future trajectory given its corresponding historical trajectory, future input series and sampled parameters. Besides, the energy function should be revised based on the assumption on identical $z$ over samples. It is worth noting that policy BNNs for the entire time sequence share the same $\mathcal{W}_s$ and $z_s$ under the sampling strategy. It indicates an underlying assumption of invariant policy over the given time horizon $h$. It enables our particle-filter-based online parameter adaptation algorithm explained later in this section. 

One problem remained is how to estimate the likelihood. We follows a simple estimation method to save computational time. Given a sample $\boldsymbol{x}_{i-m:i,s}^{n}$ at time step $i$ and corresponding sample of neural network parameters $\mathcal{W}_s$ and $z_s$, we compute the mean of predicted action $\boldsymbol{\bar{a}}^{n}_{p,i,s}$. The conditional distribution of predicted action then follows $\mathcal{N}(\boldsymbol{\bar{a}}^{n}_{p,i,s},\Sigma)$. Since the dynamics model is known, we can then estimate the conditional distribution $p(\boldsymbol{x}^n_{i+1,s}|\mathcal{W}_s,z_s,\boldsymbol{a}^n_{q,i},\boldsymbol{x}^n_{i-m:i,s})$ as $\mathcal{N}(\boldsymbol{m}^n_{i+1,s},\Sigma^n_{i+1,s})$, following the prior update procedure of optimal filtering \cite{anderson1979optimal}, e.g. Kalman filter (KF) for linear system and extended Kalman filter (EKF) for nonlinear system. A sample $\boldsymbol{x}^n_{i+1,s}$ can be computed with $\boldsymbol{\bar{a}}^{n}_{p,i,s}$ and sampled noises $(\epsilon^n_{i,s}, \omega^n_{i,s})$, along with the dynamics model for the usage of computation for later time steps. Eventually, we can approximate the likelihood function with $\prod_{i=1}^{h}\mathcal{N}(\boldsymbol{x}^{n}_{i}|\boldsymbol{m}^n_{i+1,s},\Sigma^n_{i+1,s})$.
In perspective of a single sample of neural network parameters, this estimation scheme underestimates the propagation of variance through time. However, the problem can be alleviated by increasing the number of samples $J$ to decrease the prior weight of each sample.

\begin{equation} \label{eqn:tilde_q}
\begin{aligned}
q\left(\mathcal{W},\boldsymbol{z}\right)
&=\prod_{l=1}^{L}\prod_{i=1}^{V_l}\prod_{j=1}^{V_{l-1}+1}\mathcal{N}\left(w_{ij,l}|m_{ij,l}^{w},v_{ij,l}^{w}\right)\mathcal{N}(z|m^z,v^z)^N  \\
&\propto\left[f(\mathcal{W})f(z)\right]^N p(\mathcal{W})p(z)^N,
\end{aligned}
\end{equation}
where
\begin{equation*}
\begin{aligned}
p(\boldsymbol{z})&=\mathcal{N}(z|0,\gamma),
f(z)=\exp \left\{\frac{v^z-\gamma}{2\gamma v^z}z^2+\frac{m^z}{v^z}z\right\}\\
\end{aligned}
\end{equation*}

Training the entire recurrent model has several benefits. Firstly, the distribution $q(\mathcal{W},z)$ approximates the posterior distribution conditioned on sampled trajectories. Therefore, the approximated posterior conditional distribution of trajectory inferred from the model can resemble the actual distribution better. Moreover, since the dynamics model is embedded into the recurrent model, it confines the output distribution to probability with dynamical constraints. Consequently, the parameters are optimized over dynamically feasible family of output distribution.

\subsection{Online Parameter Adaptation via Sequential Monte Carlo}
Gradient-based methods are not suitable for real-time parameter adaptation since convergence cannot be guaranteed within certain iterations. Instead, we adopt sequential Monte Carlo (SMC) sampling, namely particle filtering. SMC has been adopted for neural network training in early stage \cite{de2001sequential}. It is computationally expensive for large-scale training. In contrast, new samples collected in real time can be incorporated into posterior distribution estimation by one-step update sequentially.

Formally, we assume the weights of the policy BNN are invariant over time. Equation (\ref{eqn:model}) together with the model of policy BNN are treated as measurement model of $\mathcal{W}$ and $z$. An additional random noise ${v}_i \sim \mathcal{N}(0,\Sigma_v)$ is added in (\ref{eqn:model}). The collected $\boldsymbol{x}_{i+1}$ is then regarded as a noisy measurement of state, so that the parameters do not overfit with the new samples. To further prevent overfitting, parameters are updated each $u$ steps. At each update step, the posterior distribution of BNN parameters is estimated using particle filter. The update procedure is described in Algorithm \ref{alg:1}. For faster resampling, the weighted samples of $\mathcal{W}$ and $z$ are fitted to a Gaussian distribution, from which new samples are drawn \cite{kotecha2003gaussian}.

\section{Examples} \label{experiment}
\subsection{Toy Linear System with Stochastic Feedback Gain}

\begin{algorithm}[t] 
\caption{Parameter adaptation algorithm}
\small
\begin{algorithmic}[1] \label{alg:1}
\renewcommand{\algorithmicrequire}{\textbf{Input:}}
\renewcommand{\algorithmicensure}{\textbf{Output:}}
\REQUIRE $q_{k-u}(\mathcal{W},\boldsymbol{z})$, $\boldsymbol{x}_{k-m-u:k}$, $\boldsymbol{a}_{q,k-u:k-1}$
\ENSURE  new approximate posterior distribution $q_{k}(\mathcal{W},\boldsymbol{z})$
\STATE for $s=1,...,M$, draw samples from $q_{k-u}(\mathcal{W},z)$ and denote them as $\{\mathcal{W}_s\}_{s=1}^{M}$ and $\{z_s\}_{s=1}^{M}$.
\FOR {$s = 1,...,M$} 
    \FOR {$r = 0,...,u-1$}
        \STATE $\boldsymbol{\bar{a}}_{p,k-u+r,s}=g(\boldsymbol{x}_{k-m-u+r:k-u+r},z_s;\mathcal{W}_s)$
        \STATE approximate the likelihood by Gaussian distribution using Kalman filter or extended Kalman filter:\\ $p(\boldsymbol{x}_{k-u+r+1}|\mathcal{W}_s,z_s,\boldsymbol{a}_{q,k-u+r},\boldsymbol{x}_{k-m-u+r:k-u+r})$ $\approx \mathcal{N}(\boldsymbol{x}_{k-u+r+1};\boldsymbol{m}_{k-u+r+1,s},\Sigma_{k-u+r+1,s})$
    \ENDFOR
    \STATE $\beta_s=\displaystyle \prod_{r=0}^{u-1}\mathcal{N}(\boldsymbol{x}_{k-u+r+1};\boldsymbol{m}_{k-u+r+1,s},\Sigma_{k-u+r+1,s})$
\ENDFOR
\STATE estimate the mean and variance as: \\
\begin{equation*}
\begin{aligned}
    m^w_{ij,l}&=\frac{\sum_{s=1}^{M}\beta_s w_{ij,l,s}}{\sum_{s=1}^{M}\beta_s} \\
    v^w_{ij,l}&=\frac{\sum_{s=1}^{M}\beta_s (w_{ij,l,s}-m^w_{ij,l})^2}{\sum_{s=1}^{M}\beta_s} \\
    m^z &=\frac{\sum_{s=1}^{M}\beta_s z_s}{\sum_{s=1}^{M}\beta_s} \\
    v^z &=\frac{\sum_{s=1}^{M}\beta_s (z_s-m^z)^2}{\sum_{s=1}^{M}\beta_s} \\
\end{aligned}
\end{equation*}
\STATE obtain $q_k(\mathcal{W},\boldsymbol{z})$ based on (\ref{eqn:tilde_q})
\RETURN $q_k(\mathcal{W},\boldsymbol{z})$
\end{algorithmic}
\end{algorithm}

We start with evaluating the proposed method with a toy example, so that the predicted conditional distribution of trajectory can be compared directly with the actual distribution. The designed toy example is a linear system with stochastic feedback gain described in Equations (\ref{eqn:toy1}-\ref{eqn:toy2}).
\begin{align}
x_{i+1}&=x_i-\gamma S(\kappa_i)x_i \label{eqn:toy1}\\ 
\kappa_{i+1} &= \kappa_{i}+\zeta_{i}, \label{eqn:toy2}
\end{align}
where $\kappa_0 \sim p_1\mathcal{N}(\mu_1,v_1)+p_2\mathcal{N}(\mu_1,v_2)$ and $\zeta_i \sim \mathcal{N}(0,v_\zeta)$. The function $S(\cdot)$ is the sigmoid function. Along with the constant $\gamma$, it enables stability of the closed-loop system. A random noise $\zeta_i$ is added to have $\kappa_i$ vary over time. Given an initial state $x_0$, we want to enquire the conditional distribution $p(x_{1:15}|x_0)$ from the model. The policy BNN is expected to approximate the average gain distribution over the prediction horizon. 

\begin{figure}[t]
\centering
\includegraphics[scale=0.30]{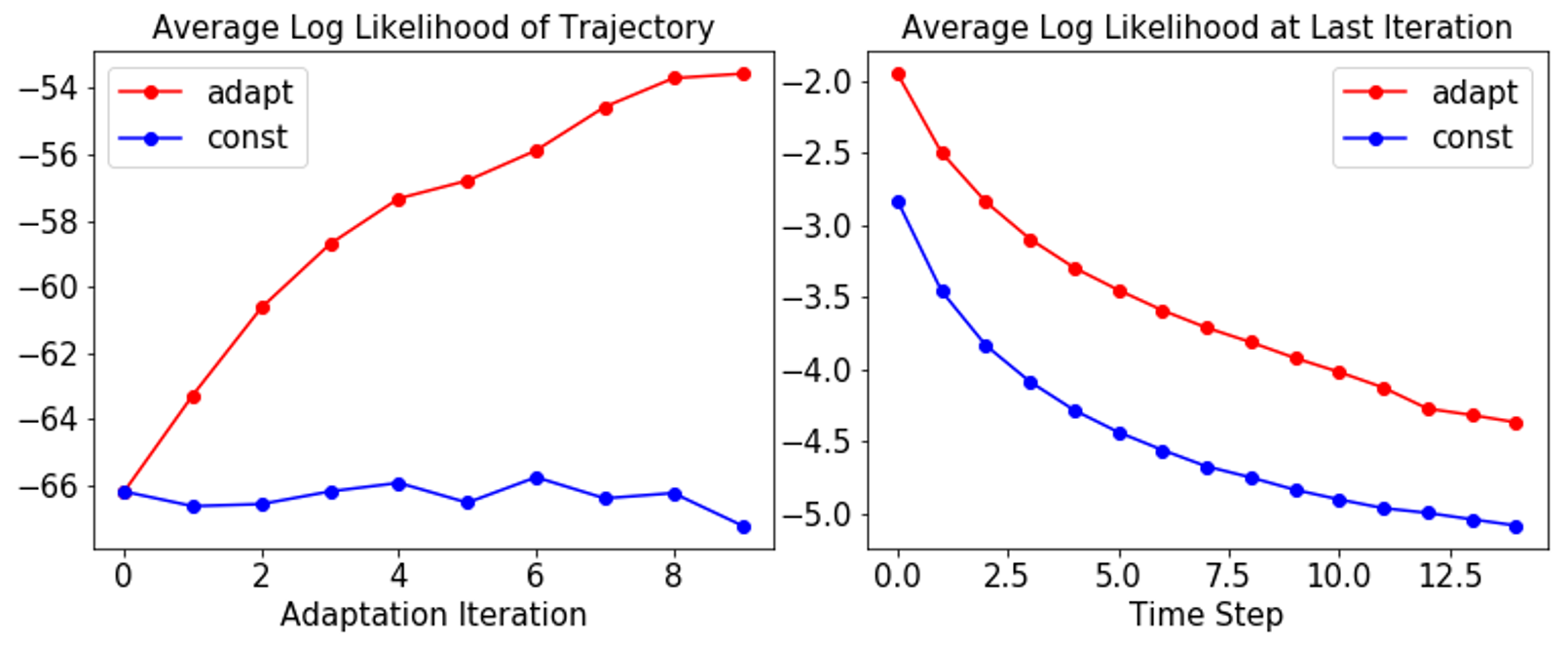}
\caption{(a) Average log likelihood over adaptation iterations. (b) Average log likelihood over the prediction horizon. Adapted model after the last iteration is compared with the original model.}
\label{fig:toy_adapt}
\end{figure}

We set $x_0=200$, $\mu_1=-1$, $\mu_2=1$, $v_1=v_2=0.36$, $p_1=p_2=0.5$ and $v_\zeta=0.04$. 5000 samples of $x_{1:15}$ are drawn for training. The network has two hidden layers with 50 hidden units per layer. The batch size is 50. The learning rate is chosen as $1e-4$. 100 samples are drawn to estimate the energy function. The model is implemented in Tensorflow \cite{abadi2016tensorflow}. We compare BNNs trained with three special $\alpha$ values, $1e-6$, $0.5$ and $1.0$. Afterwards, BRNN is trained with the best $\alpha$ value found. Moreover, a conditional GMM is trained by Expectation Maximization (EM) algorithm for comparison, so that the benefit of the complicated probability structure enabled by BNN could be illustrated. The GMM model implemented in scikit-learn \cite{pedregosa2011scikit} is adopted. 10 mixture components are selected after tuning. BNN with $\alpha=1.0$ captures the multi-modal structure of distribution, while as BNNs with other $\alpha$ values as well as GMM model fail to do so. Its performance surmounts other models in terms of log likelihood and KL-divergence as well. Training the entire recurrent model further increases the log likelihood and decreases the KL-divergence, especially for latter time steps in the prediction horizon. BRNN tends to approximate the state distribution better over the entire prediction horizon, at the cost of relatively lower similarity at the first time step.
\begin{figure*}[t]
\centering
\includegraphics[scale=0.21]{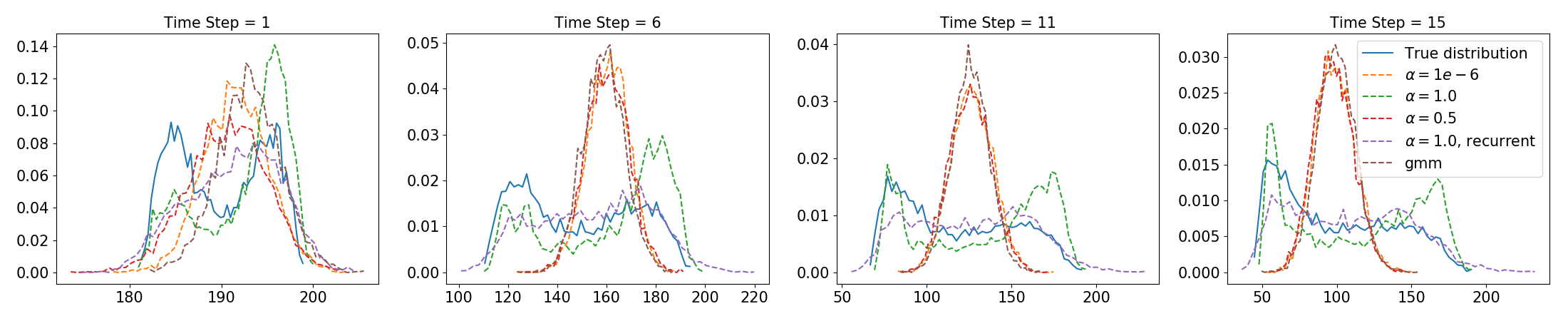}
\caption{Normalized histograms of $x_k$ at different time steps. The solid blue line represents histogram of actual $x_k$, dash lines with different color represent histogram of $x_k$ sampled from different models. The x-axis is the value of $x_k$ and the y-axis is the corresponding approximated density.} 
\label{fig:hist}
\end{figure*}
\begin{figure}[t]
\centering
\includegraphics[scale=0.27]{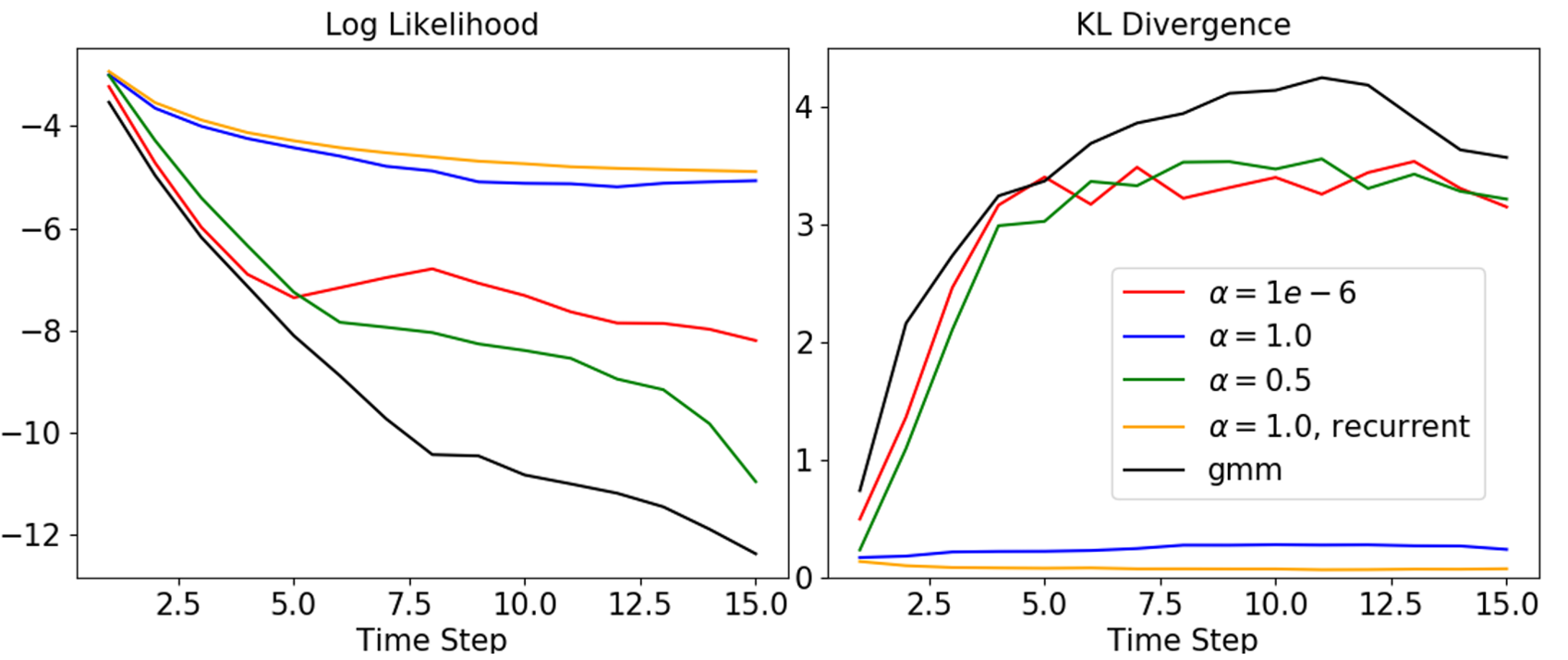}
\caption{Estimated log likelihood and KL-divergence of $x_k$ over the 15 time steps. The log likelihood is estimated as in the training stage using 100 sampled trajectories. The KL Divergence $KL(p_\text{actual}||p_\text{approx})$ is estimated based on the normalized histograms, where $p_\text{actual}$ is the actual distribution and $p_\text{approx}$ is the predicted distribution.}
\label{fig:toyplot}
\end{figure}
The online parameter adaptation algorithm is tested by adapting the policy BNN to trajectories generated with $\kappa_0\sim \mathcal{N}(\mu_1,v_1)$. 9 adaptation iterations are applied. For each iteration $i$, the policy BNN is adapted to a sampled trajectory $x^i_{0:30}$. Log likelihood of the trajectory $x^i_{0:15}$ under the predicted distribution is recorded. Meanwhile, the log likelihood computed with the unadapted model is also recorded for comparison. The test procedure is conducted for 50 times. The average log likelihood increases over iterations (Figure \ref{fig:toy_adapt}), indicating adaptation towards the distribution of new samples. 
\subsection{Interactive Car Following Model from Traffic Data}
To verify the proposed method for stochastic trajectory prediction problem in autonomous driving, we start with learning an interactive car following model from naturalistic traffic data. The scenario is stated in Section \ref{problem}. The dataset selected is Interstate 80 (I-80) Freeway Dataset from the Next Generation Simulation (NGSIM) program \cite{alexiadis2004next}. We set $\Delta t=0.2s$, $m=9$ and $h=30$. The network has three hidden layers with 50 hidden units per layer. The batch size is 250. The learning rate is chosen as $1e-4$. 1000 samples are drawn to estimate the energy function. During the training of the policy BNN, $\alpha=0.5$ achieves the best performance in terms of log likelihood. Therefore, we fix $\alpha=0.5$ and evaluate the effect of trajectory length in the training stage of the recurrent model. Particularly, we set $h=1$, $h=10$, $h=20$ and $h=30$ in (\ref{eqn:energy}) and train four models correspondingly. Meanwhile, similar to the toy example, we train a GMM policy unit for comparison. 20 mixture components are selected.

We test the trained models with 100 randomly selected trajectories from test dataset. To compare the performance in short and long time horizons separately, the average log likelihood of the first 15 time steps and the one of the latter 15 time steps are computed respectively. Increasing $h$ in training leads to higher likelihood. Since the car following scenario is relatively simple with small variance, GMM actually performs quite well especially for short time horizon. But still, BRNN with $h=30$ achieves higher log likelihood for latter time steps than GMM, indicating the effectiveness of training the entire BRNN in long-term trajectory prediction.

To test the online adaptation algorithm, we adapt the policy BNN to trajectories of the same vehicle over 90 time steps. The parameters are updated per 30 steps, i.e. 2 update iterations are applied for each trajectory. 44 trajectories are extracted from the test dataset for evaluation. Table \ref{table:car} shows the average log likelihood of the trajectories over the last 30 time steps. BRNN with $h=30$ achieves the best performance after adaptation, whereas smaller $h$ led to decreased log likelihood. To visualize the distribution for comparison, we choose a trajectory and plot the normalized histogram of predicted $v_p$ at different time steps for three models: the BRNN with $h=30$ without adaptation, the BRNN with $h=30$ after adaptation and GMM model. The variance of BRNN's prediction becomes smaller than the GMM model after adaptation, leading to larger likelihood of the actual $v_p$. It is difficult to rigorously analyze the cause as the actual distribution of data can hardly be extracted. Intuitively, the BRNN with $h=30$ learns a policy invariant over the prediction horizon, so that the adaptation algorithm based on the assumption of time-invariant policy works well. 

\begin{figure*}[t]
\centering
\includegraphics[scale=0.31]{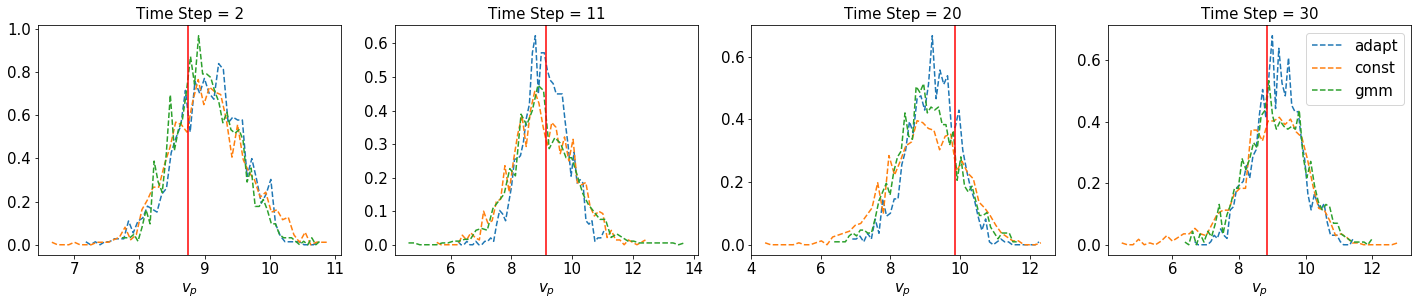}
\caption{Normalized histograms of predicted $v_p$ at different time steps. The solid red vertical line represents the actual value in the dataset. The x-axis is the value of $v_p$ and the y-axis is the corresponding approximated density.} 
\label{fig:hist2}
\end{figure*}
\begin{table*}[b]
\caption{Model Performance Comparison for Car Following Model Learning} \label{table:car}
\label{comparison}
\begin{threeparttable} 
\centering
\renewcommand\TPTminimum{\linewidth}
\makebox[\linewidth]{\scriptsize%
\begin{tabular}{|c|c|c|c|c|c|c|c|}
\hline
\multirow{2}{*}{Model}&
\multicolumn{2}{c|}{Testing Data} & \multicolumn{2}{c|}{Without Adaptation} & \multicolumn{2}{c|}{After Adaptation} \cr\cline{2-7}
 & First 15 time steps & Second 15 time steps & First 15 time steps & Second 15 time steps &  First 15 time steps &  Second 15 time steps \\ \hline
\multirow{1}{1.8cm}{BRNN, $h=1$}&
 $-3.27\pm{13.51}$ & $-4.95\pm{4.76}$ & $-4.71\pm{4.46}$ & $-7.34\pm{2.39}$ & $-191.13\pm{100.82}$ & $-40.90\pm{45.52}$ \cr \hline
\multirow{1}{1.8cm}{BRNN, $h=10$}&
 $-2.30\pm{5.50}$ & $-3.43\pm{1.15}$ & $-2.00\pm{1.05}$ & $-3.83\pm{0.52}$ & $-3.25\pm{2.04}$ & $-6.20\pm{1.31}$\cr \hline
\multirow{1}{1.8cm}{BRNN, $h=20$}&
$-2.13\pm{2.78}$ & $-3.45\pm{0.90}$ & $-2.08\pm{0.82}$ & $\mathbf{-3.52\pm{0.22}}$ & $-2.10\pm{0.81}$ & $-3.56\pm{0.32}$\cr \hline
\multirow{1}{1.8cm}{BRNN, $h=30$}&
$-2.09\pm{2.03}$ & $\mathbf{-3.39\pm{0.86}}$ & $-2.00\pm{0.73}$ & $-3.55\pm{0.32}$ & $\mathbf{-1.94\pm{0.70}}$ & $\mathbf{-3.50\pm{0.27}}$\cr \hline
\multirow{1}{1.8cm}{GMM}&
$\mathbf{-1.89\pm{2.59}}$ & $-3.46\pm{0.96}$ & $\mathbf{-1.89\pm{1.03}}$ & $-3.69\pm{0.45}$ & - & - \cr \hline
\end{tabular}}
\begin{tablenotes}
\item[1] In each cell, the average log likelihood of trajectory divided by the number of time steps is presented in form of $\text{mean}\pm{\text{std}}$. 
\item[2] $h$ is the number of future states considered at the training stage of BRNN. 
\end{tablenotes}
\end{threeparttable}
\end{table*}

\section{Discussion and Future Work} \label{discussion}
As shown in last section, the proposed BRNN architecture achieves good performance in trajectory prediction especially for long prediction horizon. Its advantages are more apparent in the toy example, where the multi-modal policy leads to complicated distribution of trajectory. The distribution in the car following scenario is relatively simple, therefore, GMM has a comparable performance with BRNN. However, BRNN still surmounts GMM in later time steps of the prediction horizon. In future research, the proposed method will be evaluated in driving scenarios where the distribution of trajectories is more complicated.

For the online adaptation algorithm, it has been illustrated in the toy example that the policy BNN can indeed adapt to online collected samples, when the assumption that the policy varies slightly over time holds. In real-world scenarios, the assumption might not hold. Based on the experimental results for car following, learning an time-invariant policy for long horizon during training seems to enable better adaptation effect, but its performance cannot be asserted. The algorithm will be further evaluated when the time-invariant assumption does not hold. Another limitation is a large number of samples has to be drawn to approximate the distribution of weights accurately when the size of BNN is large. In the future, we will investigate on reducing the number of adaptable parameters for online adaptation.

\section{Conclusion} \label{conclusion}
In this work, a novel stochastic vehicle trajectory prediction method based on BNN is developed. Combining BNN-based policy model with the known physical model recurrently, the proposed BRNN architecture can generate physically feasible trajectory distribution. Gradient-based training method enables direct optimization towards better long-term prediction performance. Furthermore, a particle-filter-based parameter adaptation algorithm is designed to adapt the policy BNN towards the driving policy or predicted target. The proposed methods achieve good performance in both the toy example with multi-modal policy and real-world car following scenario.

\addtolength{\textheight}{-0cm}   
\bibliography{reference}
\bibliographystyle{ieeetr}

\addtolength{\textheight}{+0cm}

\end{document}